\documentclass[letterpaper, 10 pt, conference]{ieeeconf}% Comment this line out if you need a4paper
\IEEEoverridecommandlockouts                              % This command is only needed if 
\usepackage{graphicx}
\usepackage{amsmath,amssymb,mathtools} 
\usepackage{multirow}
\usepackage{tabularx}
\usepackage{booktabs}
\usepackage{hyperref}

\title{\LARGE Incremental Reconstruction of Urban Environments \\ by Edge-Points Delaunay Triangulation}

\author{Andrea Romanoni$^{1}$ and Matteo Matteucci$^{2}$% <-this % stops a space
\thanks{$^{1}$DEIB, Politecnico di Milano,  Via Ponzio 34/5, 20133, Milano, Italy {\tt\small andrea.romanoni@polimi.it}  }
\thanks{$^{2}$DEIB, Politecnico di Milano,  Via Ponzio 34/5, 20133, Milano, Italy {\tt\small matteo.matteucci@polimi.it} }
}

\begin{document}

\maketitle
\thispagestyle{empty}
\pagestyle{empty}

\begin{abstract}
Urban reconstruction from a video captured by a surveying vehicle constitutes a core module of automated mapping. 
When computational power represents a limited resource and, a detailed map is not the primary goal, the reconstruction can be performed incrementally, from a monocular video,  carving a 3D Delaunay triangulation of sparse points; this allows online incremental mapping for tasks such as traversability analysis or obstacle avoidance. 
To exploit the sharp edges of urban landscape, we propose to use a Delaunay triangulation of Edge-Points, which are the 3D points corresponding to image edges. These points constrain the edges of the 3D Delaunay triangulation  to real-world edges. 
Besides the use of the Edge-Points, a second contribution of this paper is the Inverse Cone Heuristic that preemptively avoids the creation of artifacts in the reconstructed manifold surface.
We force the reconstruction of a manifold surface since it  makes it possible to apply computer graphics or photometric refinement algorithms to the output mesh.
We evaluated our approach on four real sequences of the public available KITTI dataset by comparing the incremental reconstruction against Velodyne measurements.
\end{abstract}

%----------------------------------------------------------------------------------------------------------------------------
\section{Introduction}
\label{sec:intro}
Urban 3D reconstruction represents a fundamental task of many robotics applications, e.g, city mapping \cite{Pollefeys_et_al_08} or city segmentation \cite{Hane_et_al_09} from a surveying vehicle.
Most of the existing systems propose computationally expensive stereo methods that build a very detailed reconstruction by estimating dense keyframe depth maps , usually by means of GPU computing \cite{Pollefeys_et_al_08,Cornelis_et_al08}. 
However, in some robotics applications, a monocular, rough and computationally less expensive reconstruction is preferred,  for instance, let consider traversability analysis performed on embedded CPU-only systems deployed with a single camera.

Space carving \cite{Seitz_et_al06} thus becomes an effective method to build  a large urban map quickly. 
It usually bootstraps from a sparse point cloud, estimated, for instance, through Structure from Motion \cite{Snavely_et_al06}. 
Out of this sparse point cloud space carving builds a convenient partition of the space, usually a 3D Delaunay triangulation, where each part, e.g., the tetrahedron, is initialized as \emph{matter}.
Then, a ray tracing procedure marks as \emph{free space} the parts  crossed by a camera-to-point \emph{viewing ray}, i.e., the segment from a camera center to a 3D point in the triangulation. 

Existing literature proposes both batch \cite{Pan_et_al09} and incremental \cite{Litvinov_Lhuillier_13,Lovi_et_al_11} space carving methods.
The former perform the reconstruction by taking into account all the viewing rays at the same time; the latter carve the space incrementally, i.e., frame-by-frame.
In our case we focus on the incremental approach, since we address the scenario of a surveying vehicle that builds its own map of the city while navigating through it.

The authors in \cite{Lovi_et_al_11}  and \cite{Litvinov_Lhuillier_13} propose two incremental space carving algorithms based on the 3D Delaunay triangulation of sparse 3D point clouds. 
In \cite{Lovi_et_al_11} the estimated surface is simply the boundary between free space and matter; on the other hand in \cite{Litvinov_Lhuillier_13}, and its extension \cite{litvinov_Lhiuller14}, the estimated surface is forced to be \emph{manifold}, i.e., for each vertex, the neighboring triangles are homeomorphic to a disk. 

Several reasons lead to enforce the manifold property. First, most computer graphics algorithms need the manifold property to hold, one example is the Laplace-Beltrami operator \cite{Meyer03}.
Moreover, photometric surface refinement as in \cite{vu_et_al_2012} and \cite{Delaunoy08} usually needs surface manifoldness to properly compute the gradient flow that minimizes the photometric error. Finally, non-manifold surfaces are usually not realistic in real world environments.

\begin{figure}
\centering
\begin{tabular}{cc}
\includegraphics[width=0.75\columnwidth]{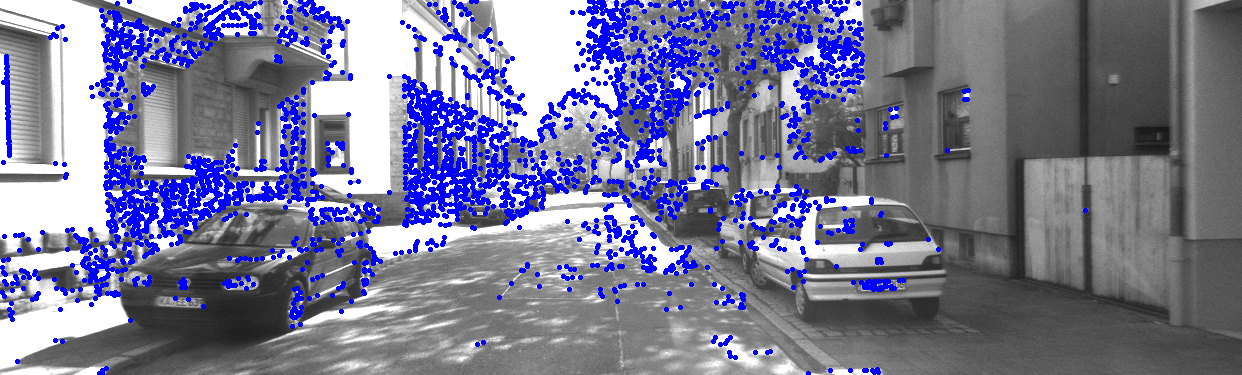}\\
(a)\\
\includegraphics[width=0.75\columnwidth]{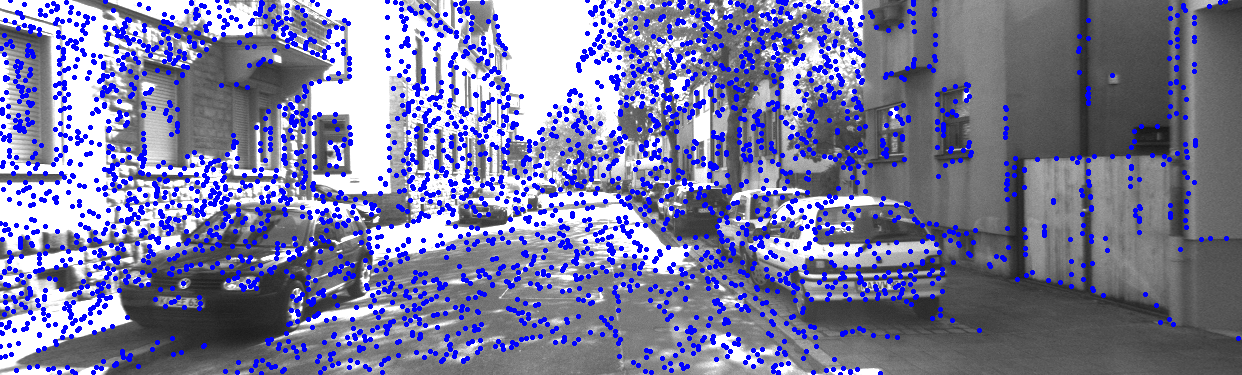}\\
(b)\\
\end{tabular}
\caption{Different features extracted on the same image: (a) shows 3609 Harris corners, (b) shows 3595 Edge-Points.}
\label{fig:Edge-Points}
\end{figure}

\begin{figure}
\centering
\begin{tabular}{cc}
\includegraphics[width=0.75\columnwidth]{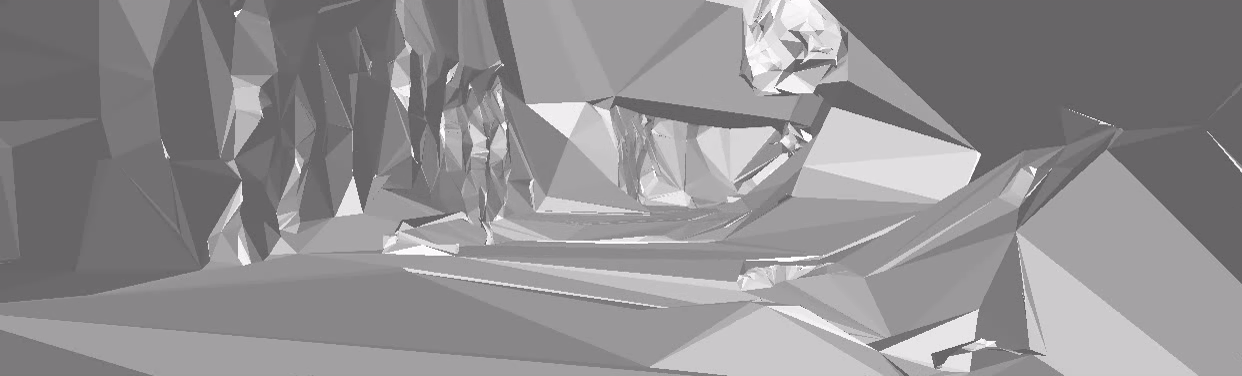}\\
(a)\\
\includegraphics[width=0.75\columnwidth]{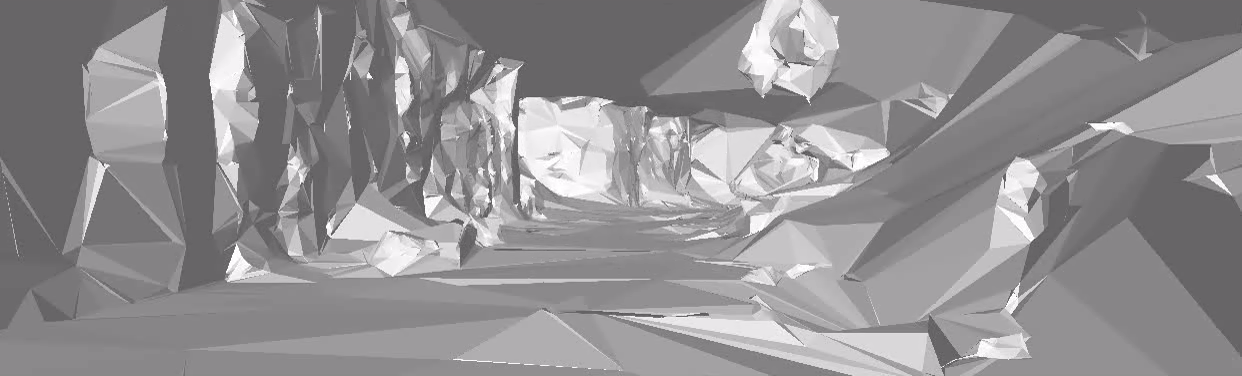}\\
(b)\\
\includegraphics[width=0.75\columnwidth]{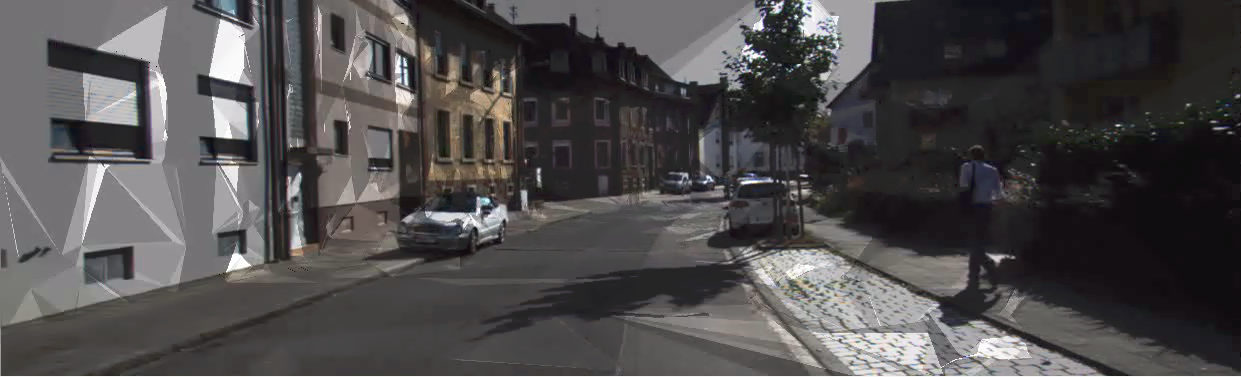}\\
(c)\\
\end{tabular}
\caption{Reconstructions with (a) Harris corners, (b) Edge-Points, and (c) textured reconstruction.}
\label{fig:recons}
\end{figure}

In this paper, we improve on the approach of \cite{Litvinov_Lhuillier_13} to extract a manifold incrementally. 
Differently from \cite{Litvinov_Lhuillier_13}, instead of reconstructing 3D points corresponding to Harris features, we propose to build the 3D Delaunay triangulation on the points projecting on the (Canny) edges of the images, named \emph{Edge-Points} (see Fig. \ref{fig:Edge-Points}). 
The existing incremental space carving systems, e.g., \cite{litvinov_Lhiuller14, Litvinov_Lhuillier_13, Lovi_et_al_11}, rely on sparse point cloud estimated by Structure from Motion, which discards the Edge-Points.

The main drawback of these points is the degree of freedom along the edge itself that usually causes instability in estimation and matching.
Nevertheless, several reasons supports the usage of Edge-Points.
First, urban scenarios show lot of sharp edges, therefore Edge-Points represent suitable vertexes to constrain the edges of the 3D Delaunay triangulation to real-world edges (see Fig. \ref{fig:recons}). 
Then, as Fig. \ref{fig:Edge-Points} shows, Edge-Points provide a better coverage of the image.
Finally, the number of Edge-Points is easier to tune with respect to the classical feature detector: by changing the downsampling rate we change proportionally the number of Edge-Points.

Other authors \cite{Rhein_et_al13, Tomono09} already took advantage of Edge-Points in their systems. Rhein et al. in \cite{Rhein_et_al13} propose a heterogeneous (corner features and Edge-Points) tracker that exploits the epipolar constraint, but, differently from us, their work is focused on the tracking stage and it aims at a sparse point cloud reconstruction.
Tomono \cite{Tomono09}  uses Edge-Points to make  the Simultaneous Localization And Mapping (SLAM) process robust in an indoor scenario that exhibits a lack of texture, but he does not reconstruct the 3D surface of the scene, moreover it uses a stereo rig that makes the estimation of the 3D point positions easier with respect to the monocular case addressed in this paper.

The use of a monocular camera looking forward induces low parallax which makes  the estimation of the 3D point positions not a trivial task. This issue adds to the previously mentioned Edge-Points instability. 
In this paper we show that the combination of the  Kanade-Lucas-Tomasi (KLT) tracker \cite{Lucas_Kanade81}, a convenient filtering of the matches, and Gaussian-Newton optimization successfully handle Edge-Points estimation even in this complex scenario. 

Our system bootstraps from a good estimate of the camera poses, as in many urban 3D reconstruction systems \cite{ Pollefeys_et_al_08, Cornelis_et_al08}. 
We assume the camera pose estimation could be obtained with a Structure from Motion or SLAM technique, e.g., \cite{Snavely_et_al06},  or with an estimate obtained by fusing GPS, inertial sensors and visual odometry such as in \cite{Cucci_Matteucci14}.
This assumption enables us to estimate independently, and very efficiently, the 3D Edge-Point positions.

Besides the use of Edge-Points, a second contribution in this paper addresses the visual artifacts issue that sometimes affects the estimated surface.
This issue was deeply studied in \cite{litvinov_Lhiuller14}, where the authors propose an ad hoc post-processing procedure that attempts to detect and remove the artifacts by preserving the manifold property; it runs quite fast, but its computational complexity is not negligible ($0.43$s per-frame).  
In this paper we propose a very efficient heuristic to preemptively avoid visual artifacts, which runs significantly faster than the previously mentioned procedure (around $0.001$-$0.010$s per-frame). We named this heuristic Inverse Cone Heuristic since the space affected by the heuristic has the shape of a cone directed inversely with respect to camera-to-point viewing ray.

In Section \ref{sec:manifold} we summarize the approach of \cite{Litvinov_Lhuillier_13} to reconstruct a manifold surface.
In Section \ref{sec:3D-Reconstruction} we describe our incremental reconstruction system, focusing on the 3D Edge-Point cloud estimation and the preemptive approach to remove the visual artifacts. 
In Section \ref{sec:experimental-results} we show the experimental results on the public available dataset KITTI \cite{Geiger_et_al12}, while in Section \ref{sec:conclusion} we conclude the paper.

%----------------------------------------------------------------------------------------------------------------------------
\section{Manifold Reconstruction}%----------------------------------------------------------------------------------
%\subsection{The manifold property}
\label{sec:manifold}
We are interested in reconstructing a manifold surface in the 3D space which represents the observed scene.
A surface is manifold if and only if the surface neighborhood of each point is homeomorphic to a disk.
In the discrete case, the points are the vertexes of the mesh, while the incident triangles (or polygons) form the neighborhood. 
So, a discrete surface is manifold if each vertex $v$ is \emph{regular}, i.e., if and only if the edges opposite to $v$ form a closed path without loops \cite{lhuillier_Yu2013}. 

\subsection{Incremental manifold extraction}
\label{subsec:incrementalManifold}
In this section we briefly summarize the method introduced in \cite{Litvinov_Lhuillier_13} and \cite{litvinov_Lhiuller14}, which, in this paper, we enhance significantly by choosing a proper point cloud to build the Delaunay triangulation and by avoiding most  of the artifacts in the estimated surface with the Inverse Cone Heuristic.

In \cite{Litvinov_Lhuillier_13}, the authors bootstrap the surface reconstruction from a manifold by partitioning the 3D triangulation of the space between the set $O$ of \emph{outside} tetrahedra, i.e., the manifold subset of the free space (not all free space tetrahedra would be included in this set), and the complementary set $I$ of \emph{inside} tetrahedra, i.e., the remaining tetrahedra that roughly represent the matter.

Let  $\delta( O_{t_{\text{init}}})$ be the initial manifold, i.e., the boundary between $O$ and $I$, estimated as it follows:
\begin{itemize}
  \item \emph{Point Insertion}: add all the 3D points estimated up to time $t_{\text{init}}$ and construct their 3D Delaunay triangulation;
  \item \emph{Ray Tracing}: mark the tetrahedra as free space according to the viewing rays: the list of these tetrahedra is named $F_{t_{\text{init}}}$;
  \item \emph{Growing}: initialize a queue $Q$ with the tetrahedron $\Delta_1 \in F_{t_{\text{init}}}$ that gets the highest number of viewing ray intersections; then iterate the following procedure until $Q$ is empty: (a) remove the tetrahedron $\Delta_{\text{curr}}$ with the highest number of viewing ray intersections from $Q$; (b) add it to $O_{t_{\text{init}}}$ only if the resulting $\delta (O_{t_{\text{init}}} \cap \Delta_{\text{curr}})$  is manifold; (c) add to the queue $Q$ its neighboring tetrahedra that are not already inside the $O_{t_{\text{init}}}$ set. 
\end{itemize}

Once the system is initialized, a new set of points $P_{t_k}$ is generated at $t_k= t_{\text{init}} + k*T_k$, where $k \in \mathbb{N^+}$ is the keyframe index and $T_k$ is the period. 
The insertion of each point $p\in P_{t_k}$ would cause the removal of a set $D_{t_k}$ of the tetrahedra that invalidates the Delaunay property; the surface $\delta (O_{t_k}) = \delta (O_{t_{k-1}} \setminus D_{t_k})$ is not guaranteed to be manifold anymore. 
To avoid this, the authors in \cite{Litvinov_Lhuillier_13} define a new list of tetrahedra $E_{t_k} \supset D_{t_k}$ and apply the so called \emph{Shrinking} procedure, i.e., the inverse of Growing: they subtract iteratively form $O_{t_{k-1}}$ the tetrahedra  $\Delta \in E_{t_k}$ keeping the manifoldness valid.
After this process, it is likely that $D_{t_k} \cap O_{t_k} = \emptyset$; however, in the case of $D_{t_k} \cap O_{t_k} \neq \emptyset$ the point $p$ is not added to the triangulation.
Once all points in $P_{t_k}$ have been added (or dropped), the \emph{growing} process runs similarly to the initialization procedure, but the queue $Q$ is initialized with the tetrahedra $\Delta \in T \setminus O$ such that  $\Delta \cap \delta O \neq \emptyset$.

%----------------------------------------------------------------------------------------------------------------------------

\section{3D Reconstruction with Edge-Point and Inverse Cone Heuristic}
\label{sec:3D-Reconstruction}
3D reconstruction with space carving entails space discretization.
We choose the 3D Delaunay triangulation to partition the space into tetrahedra since it has been recognized in the literature to be a convenient representation for scene reconstruction \cite{Litvinov_Lhuillier_13, Pan_et_al09, labatut2007efficient, Lovi_et_al_11}.
In the following we show how we choose and we estimate the sparse point cloud upon which the triangulation is created and how we conveniently carve the space to preemptively avoid artifacts in a novel way that deeply differs from the approach of \cite{litvinov_Lhiuller14}.

%----------------------------------------------------------------------------------------------------------------------------
\subsection{Edge-Points for 3D Delaunay triangulation}
\label{subsec:pcl_estimation}
A key aspect for a 3D reconstruction pipeline, not stressed enough in the literature, is the choice of the points on which the Delaunay triangulation is built, i.e., what kind of points made up the sparse point cloud.

As most of man-made environments, urban scenarios show a lot of sharp edges, e.g., the corners of building fa\c{c}ades, the borders of windows, or the silhouettes of parked cars. Existing 3D space carving systems do not leverage on this information, but they rely only on maximally stable points.
Stable points are suitable features to track, however, a reconstruction relying on them over-simplifies the reconstructed world: it mostly fails to capture the sharp edges which do not show sharp corners too.

\begin{figure}
\begin{center}
\begin{tabular}{cc}
\centering
\includegraphics[width=0.3\columnwidth]{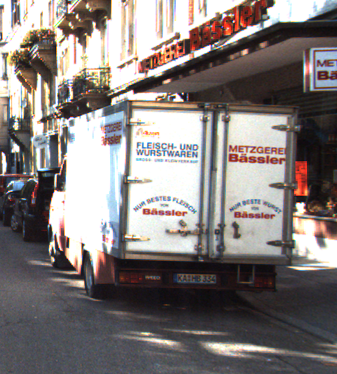}&
\includegraphics[width=0.3\columnwidth]{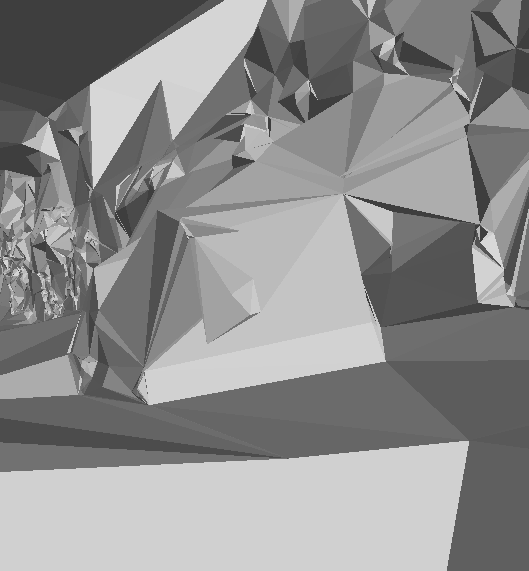}\\
(a) & (b)\\
\includegraphics[width=0.3\columnwidth]{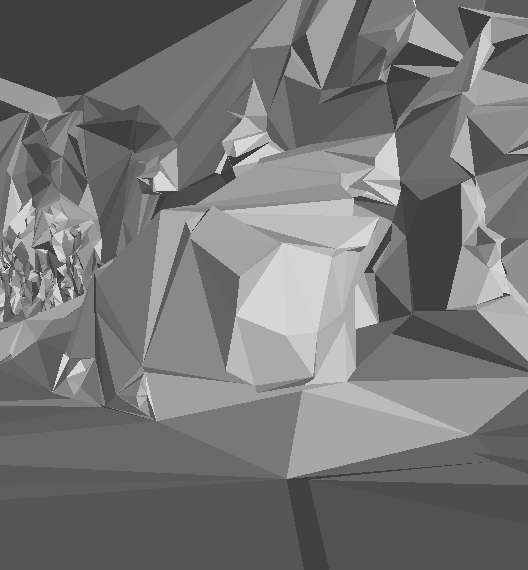}&
\includegraphics[width=0.3\columnwidth]{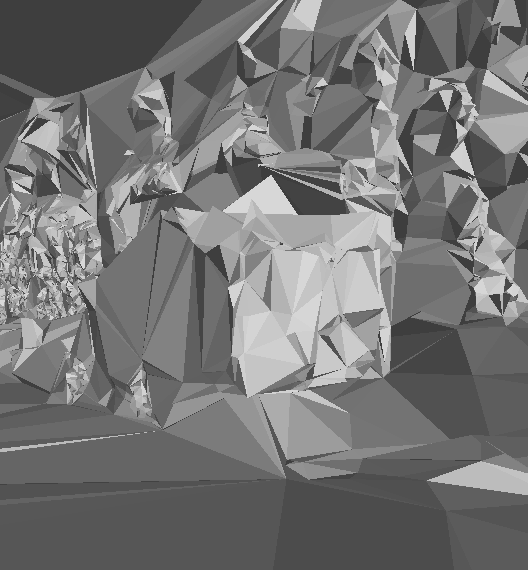}\\
(c) & (d)\\
\end{tabular}
\end{center}
\caption{Three examples of reconstruction from sparse data of the light truck in (a): with Harris corners in (b), with Edge-Points with downsample rate  $\frac{1}{40}$  in (c)  and Edge-Points with  downsample rate  $\frac{1}{10}$ in (d).}
\label{fig:reconstrEx}
\end{figure}

We propose to overcome this limitation by estimating the 3D position of Edge-Points, so to constrain the edges of the 3D triangulation to lay close to \emph{real-world} 3D edges. 
This novel triangulation generates a carved space more faithful to the real-world scene, see for instance the difference of the truck of Fig. \ref{fig:reconstrEx}(a) reconstructed with the use of Harris corners in Fig. \ref{fig:reconstrEx}(b), and with the Edge-Points at a low ($\frac{1}{40}$) and high ($\frac{1}{10}$) downsampling rate  in Fig. \ref{fig:reconstrEx}(c) and \ref{fig:reconstrEx}(d) respectively. 

Fig. \ref{fig:Edge-Points} shows that some Edge-Points have been induced by grass and shadows on the road plane. Even if this subset of features will not lay on real-world edges, their presence does not affect the quality of the reconstruction, since they lay, or are close, to the actual \emph{matter}.
Nevertheless most of these points, especially those induced by the grass, get extracted also by the Harris corner detector. 
Let stress one more time here that most of the Edge-Points lay on the real-world edges, while the Harris corners do not locate them very well.
Another important point is that is easy to increment the number of Edge-Points used to reconstruct the environment by keeping the feature quality unchanged, while the corner-like features quality usually degrades as soon as other features are required. 
To verify this, in the experimental section we tested our algorithm with two different downsampling rates showing that the quality of the reconstruction significantly improves as the sampling rate increases.

%----------------------------------------------------------------------------------------------------------------------------

\subsection{Edge-Points tracking and reconstruction}
\label{subsubsec:Edge-Point-tracking}
\begin{figure}[t]
\centering
\includegraphics[width=0.8\columnwidth]{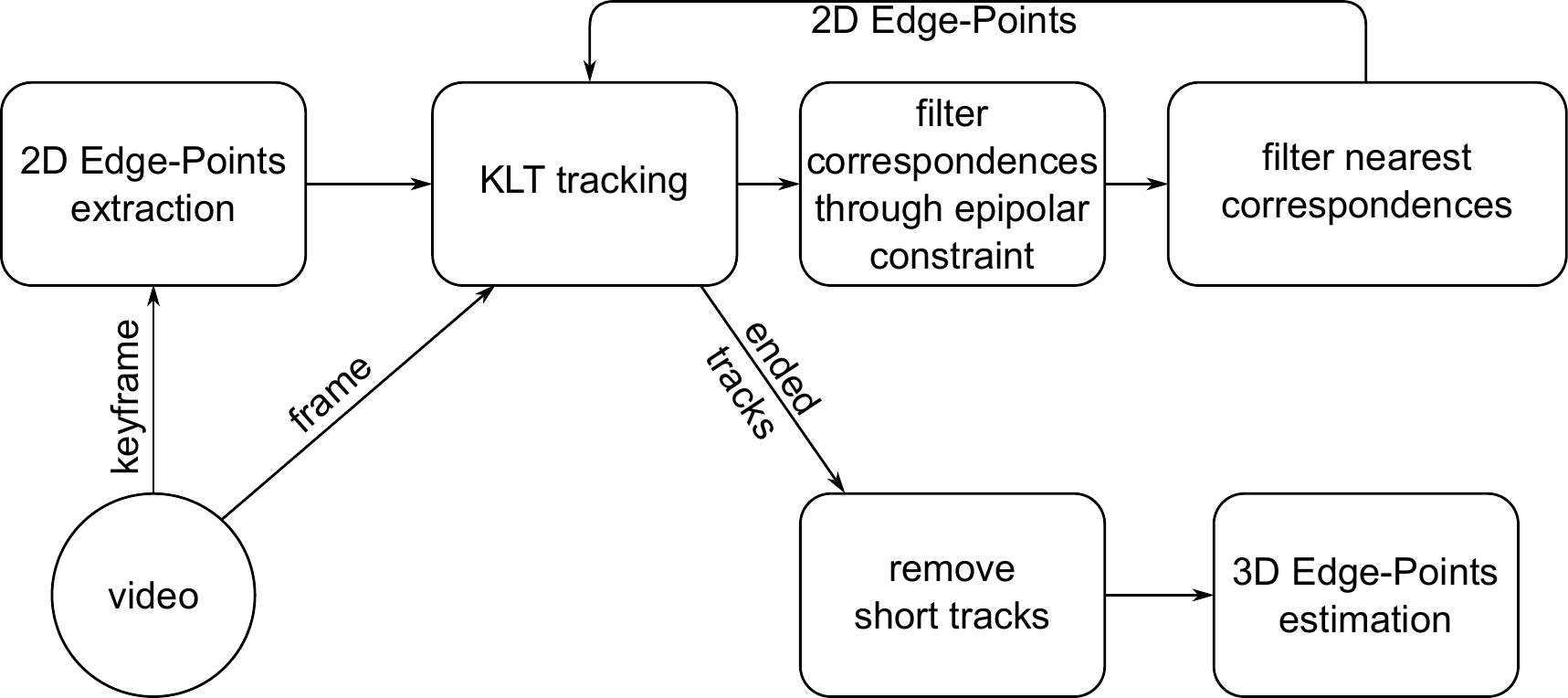}
\caption{Edge-Point tracking and estimation process.}
\label{fig:algorithm}
\end{figure}

In Fig. \ref{fig:algorithm} we depict the tracking and estimation process.
For each \emph{keyframe}, i.e., one frame every $T_k$, we extract the 2D Edge-Points by (a) estimating the image edges with the Canny algorithm, and (b) downsampling those edges with step $T_{edges}$, i.e., we downsample the chains of pixels that made up the edges. 
Then we track these points in consecutive frames (both in keyframes and non-keyframes). Each \emph{track}, i.e., the sequence of point 2D positions in subsequent images, contains the \emph{measurements} of a 3D point. The value of $T_k$ depends on the camera speed; in our case of surveying vehicle, is fixed such that we have two keyframe per-second.

We track the 2D Edge-Points with the KLT tracker \cite{Lucas_Kanade81} as suggested by Rhein et al. \cite{Rhein_et_al13} because it enables faster reconstruction with respect to more complex trackers which, for instance, relies on SIFT descriptor computation.
KLT tracks successfully most of the 2D Edge-Points between two consecutive frames; however, to reach good 3D point position estimates, we need to take into account errors due to the low parallax induced by the forward motion of the monocular camera and we need to filter out wrong correspondences produced by the mentioned edge instability.

The low parallax issue affects the estimation process when the camera looks towards the moving direction. The uncertainty of the 2D point measurement on the image plane is usually assumed to be Gaussian, so the measurement uncertainty spreads in the 3D space, through the uncertainty ellipse on the image plane, as a cone whose vertex is in the camera center. 
When the parallax is low, the uncertainty cones of consecutive measurements of a 3D point are almost overlapped \cite{HaZi04}. As the intersection of these uncertainty cone becomes relevant, the 3D point position estimation is no more reliable.

To ensure  an overall significant parallax and to successfully estimate the 3D points position, we filter the tracks  both at a local and at a global level.
At a local level we filter out an Edge-Point when the displacement of its two consecutive measurements is too small, i.e., when the parallax is almost null and the uncertainty of its estimate tends to infinite. 
We experimentally set this minimum displacement to $d_{meas}^{min} = 5 \text{pixels}$ on our videos, but this parameter is related to the video frame rate and the camera focal length so it should be adapted according to the specific setup. 
Usually, as expected from the forward motion, the points filtered out lay around the center of the image, where the parallax is lower.

At a global level we discard also short tracks, i.e., those tracks containing less then $l_{track}^{min}$ measurements (we recall here that a track is the sequence of measurements of a 3D Edge-Point in subsequent images), where in our setup $l_{track}^{min} = T_K$ (recall that $T_K$ is the keyframe period).

To add robustness to the tracking step and manage the well-known instability  of Edge-Points  we drop the correspondences that do not satisfy the epipolar constraint. 
Let $x_{t-1}$ and $x_t$ be two corresponding points in frame $t-1$ and $t$, and $F$ the fundamental matrix between the camera at time $t-1$ and $t$, the following equation holds:
\[
 x_{t-1}^{T}Fx_t = 0 .
\]

Given $K_{t-1}$ and $K_t$ the intrinsic parameters of the two cameras, which are the same in the monocular case, and assuming the world reference frame fixed in camera $t-1$ and $P = [R_t,t_t]$ to be the pose of camera $t$, the fundamental matrix can be computed as:
\[
F = K_t^{-T}R_tK_{t-1}^T [K_{t-1}R_{t}t_t]_x
\]
where $[.]_x$ is the skew-symmetric operator \cite{HaZi04}.
Given a point $x_{t-1}$ and the matrix $F$, the vector $l_{t} = Fx_{t-1}$ is the epipolar line, i.e., the locus of the points corresponding to $x_{t-1}$ in the image plane of camera $t$. 

The epipolar constraint states that, given a point $x_{t-1}$, the corresponding point $x_{t}$ lies on the epipolar line. So, given the KLT correspondence $x_{t-1}$-${x}_{t}$, we drop it if $\text{dist}_{L_2}\{Fx_{t-1}, {x}_{t}\} <\epsilon_e$, where $\epsilon_e$ is fixed to a tolerant value of $\epsilon_e = 20px$ due to the noise of the epipolar contraint estimation induced by the forward motion.
The epipolar constraint represents a necessary, but not sufficient condition. The remaining wrong correspondences will be filtered during the 3D point estimation step.

The previous filtering approach is intended to deal with almost-static scene so it filters out most of  the dynamic objects. This behavior is especially suitable for mapping purposes: in this case the map usually has not to include dynamic object such as moving cars or pedestrians. 

%----------------------------------------------------------------------------------------------------------------------------
\subsubsection*{Parallel 3D Point Positions Estimation}
\label{subsubsec:3D-point-estimation}
Several space carving reconstruction algorithms adopt a Structure from Motion technique to estimate both  the camera pose and the 3D points position at the same time, see for instance \cite{Yu_Lhuillier12, Litvinov_Lhuillier_13} and \cite{Lovi_et_al_11}. 
On the other hand, in urban applications, especially those involving autonomous vehicles, a very good estimate of the camera pose can be derived from sensors that are different from the camera itself (for instance with the sensor fusion technique in \cite{Cucci_Matteucci14}). 
Therefore, as in  many urban reconstruction systems (\cite{ Pollefeys_et_al_08, Cornelis_et_al08})  we assume the camera pose to be known, while triangulating the 3D edge-points. This assumption allows to estimate the 3D position of each 3D Edge-Point independently, i.e., in parallel.

After the tracking process, for each Edge-Point we first estimate a rough 3D position by triangulating the first and last measurements with the classic algorithm proposed by Hartley and Sturm \cite{Hartley_Sturm97}. 
We then optimize this 3D position estimate with a Gauss-Newton algorithm by minimizing the 3D reprojection error over the whole track (we fixed a number of $N_{GN} = 50$ iterations):
\begin{equation}
 e(X_{3D})^i = P^i \cdot X_{3D} - x_{meas}^i, \forall i \in \text{track}
\end{equation}
where $P^i$ is the $i$-th camera matrix, $X_{3D}$ is the 3D position of the point to be estimated, and $x_{meas}^i$ is the  measurement in the $i$-th image.
Since some wrong correspondence could exist, we drop the Edge-Points for which the mean reprojection error is higher then $\epsilon_{GN} = 2\text{px}$ at the end of optimization.

% 
% \begin{figure}
% \centering
% \begin{tabular}{cc}
% \includegraphics[width=0.45\columnwidth]{./conicCarvingContinue}&
% \includegraphics[width=0.45\columnwidth]{./conicCarvingImplemented}\\
% (a) & (b)
% \end{tabular}
% \caption{The inverse cone heuristic applied to the ray tracing step. On the left the inverse cone, and on the right our implementation in the Delaunay triangulation domain. Darker blue corresponds to higher weights.}
% \label{fig:ConicCarving}
% \end{figure}

\subsection{Inverse Cone Heuristic for preemptive visual artifacts removal.}
\label{sec:visualartifacts}
\begin{figure}
\centering
\begin{tabular}{cc}
\includegraphics[width=0.35\columnwidth]{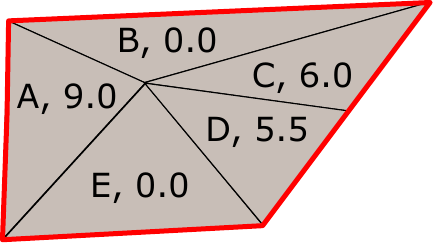}&
\includegraphics[width=0.35\columnwidth]{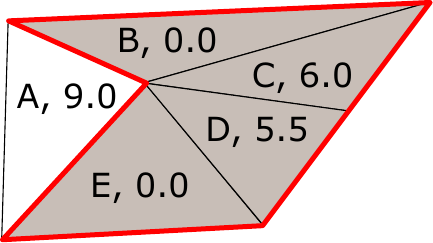}\\
(a) & (b)
\end{tabular}
\caption{Simple scenario where a visual artifact is generated. Each triangle is labelled as ``name, weight''. The bold red line is the boundary between inside (dark triangles) and outside (white triangles) sets. Bootstrapping from (a), the triangle A is added to the manifold in (b), then neither C or D can be added anymore without invalidating the manifold property.}
\label{fig:artifact}
\end{figure}

%----------------------------------------------------------------------------------------------------------------------------
Once the 3D Edge-Points have been estimated, we propose an enhanced version of the algorithm in \cite{Litvinov_Lhuillier_13} to reconstruct a manifold surface.
The surface extracted with the algorithm described in Section \ref{sec:manifold} is affected by visual artifacts.
Fig. \ref{fig:artifact} shows a simple scenario where a visual artifacts is generated due to the order of tetrahedra addition to the manifold.
The algorithm bootstraps from the manifold in Fig. \ref{fig:artifact}(a); then it grows the manifold by we add the triangle A (Fig. \ref{fig:artifact}(b)). Afterward, C and D are free space but they are kept in the inside set, otherwise they invalidate the manifold property; this two triangles make up a visual artifact.

In our case the visual edges which are critical for the reconstruction quality are those containing at least one edge long enough to be considered unrealistic. 
More formally, Litvinov and Lhuiller, in \cite{litvinov_Lhiuller14}, define a \emph{critical visual artifact} as a set of tetrahedra belonging to the free space, but not included in the outside set and which contains at least one \emph{visually critical edge}, i.e., an edge $ab$ such that exists a camera center $c$ such as $\widehat{acb}>\alpha$, where $\alpha$ is a user defined parameter (in our algorithm we do not need to define this parameter).

\begin{figure}
\centering
\begin{tabular}{cc}
\includegraphics[width=0.35\columnwidth]{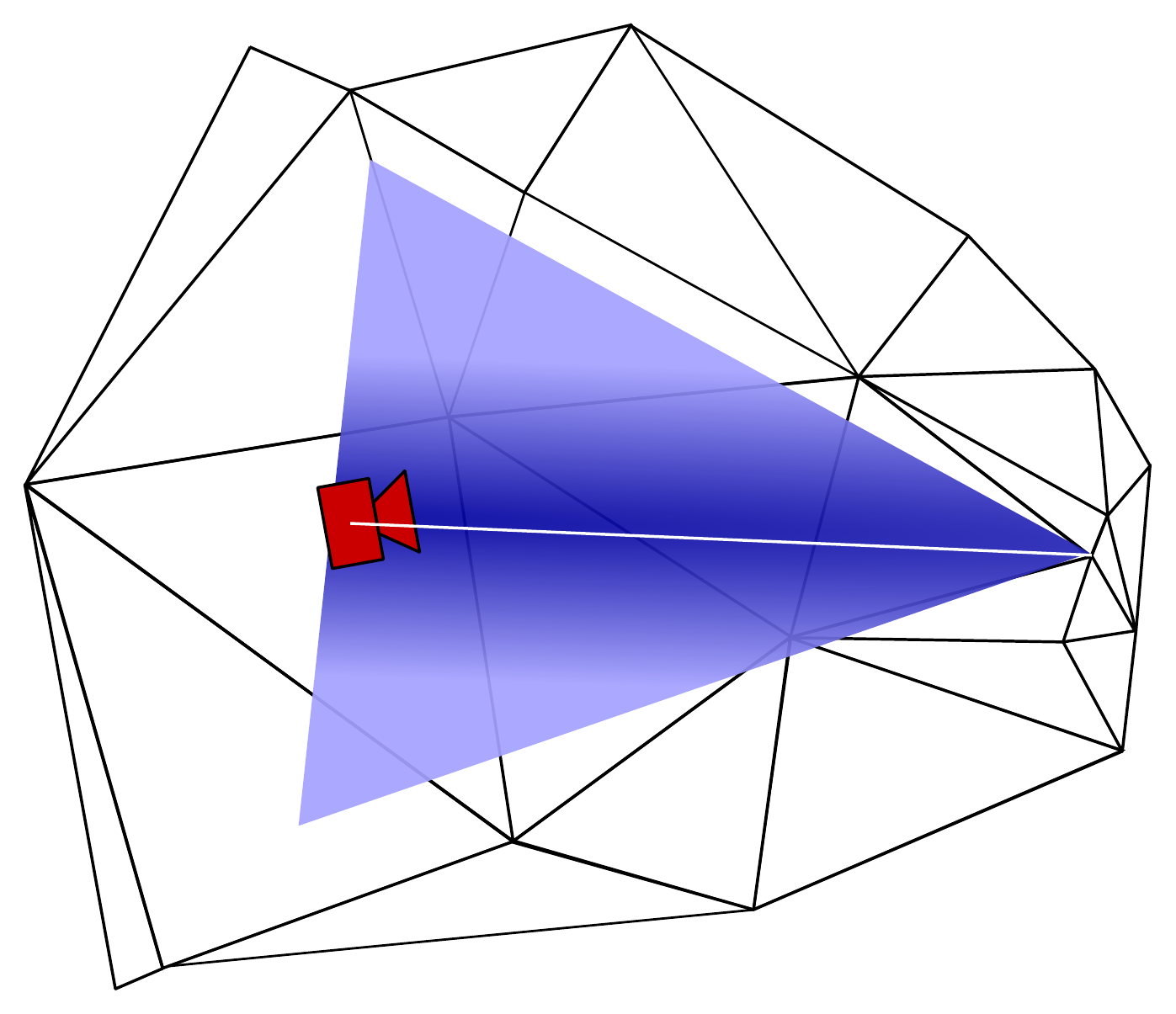}&
\includegraphics[width=0.35\columnwidth]{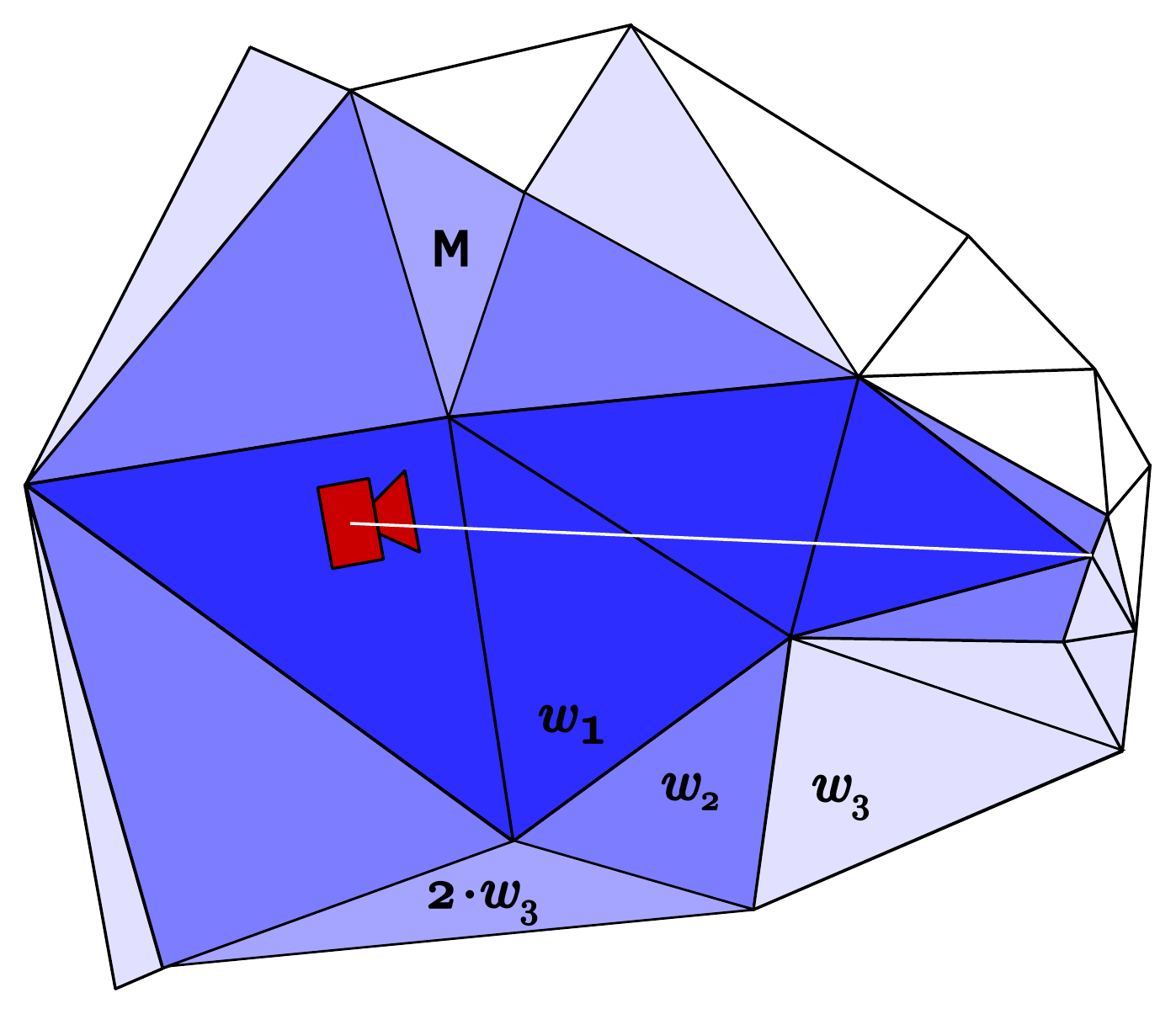}\\
(a) & (b)
\end{tabular}
\caption{The inverse cone heuristic applied to the ray tracing step. On the left the inverse cone, and on the right our implementation in the Delaunay triangulation domain. Darker blue corresponds to higher weights.}
\label{fig:ConicCarving}
\end{figure}

Litvinov and Lhuiller in \cite{litvinov_Lhiuller14}, propose a post-processing method to detect and remove critical artifacts keeping the manifold property valid.
In this paper, we propose a preemptive approach significantly different, and complementary, with respect to \cite{litvinov_Lhiuller14}. 
Our idea relies on two observations: the tetrahedra that likely turn into visually critical edges are big tetrahedra since they contains long edges, and big tetrahedra are mostly close to the camera path.

As the example of Fig. \ref{fig:artifact} shows, the order of growing is a key point to avoid the creation of visual artifacts; thus,
 by modifying the ray tracing step, we aim preemptively enforce a carving order such that big tetrahedra near to the camera become the first to be added to the reconstructed manifold.
We replace the intersection count associated to each tetrahedron with a weight. Ideally in the continuous space, we would apply  a cone-shaped weighting heuristic, we named Inverse Cone Heuristic, which opens inversely with respect to the ray sense (see Fig. \ref{fig:ConicCarving}(a)),  such that the region receiving weights increments gets smaller and smaller as the ray approach the viewed point.
In the real discrete implementation, for each ray from the camera to the 3D point, we increment by $w_1$ the weights of the traversed tetrahedra, by $w_2$ the weights of their neighbors and by $w_3$ the weights of the neighbors of the latter tetrahedra.
Since big tetrahedra are near to the camera, this induces the cone-shaped weighting scheme as in Fig. \ref{fig:ConicCarving}(b).

As Fig. \ref{fig:ConicCarving}(b) shows, some ``neighbors of neighbor'' tetrahedra receive more than one increment, in particular they receive up to 4 multiple increments (one for each neighbor), but in practical cases they are usually 2 or 3.
Multiple increments let to spread the weights to tetrahedra close to the ray, without any  neighboring facet to the traversed tetrahedra, e.g.,  triangle M in Fig. \ref{fig:ConicCarving}(b).
To avoid high weights due to multiple increments, we tune the value of $w_3$ such that the maximum increment for neighbors of neighbor is equal to $w_2$.
For all the datasets we fixed $w_1$ to a reference value of $1.0$, then $w_2$ to a close value of $0.8$ and $w_3 = \frac{w_2}{4} = 0.2$, where 4 represents the maximum number of multiple increments received by one tetrahedra for a single ray.

After the ray tracing step, the region growing and shrinking procedures follows the ordering induced by the computed weights, but, to avoid carving the actual matter, one tetrahedron is added, or subtracted, to the manifold only if is traversed by at least one camera-to-point ray.

\section{Experimental results}
\label{sec:experimental-results}
A monocular 3D reconstruction benchmark for urban scenarios, with accurate ground truth, does not exist; then we evaluated our contribution on four different sequences of the public available dataset \cite{Geiger_et_al12}. 
This dataset contains a Velodyne HDL-64E point cloud for each sequence which can be used as ground truth for 3D reconstruction validation.
The video stream was captured by a Point Grey Flea 2 camera, which took $1392\text{x}512$ gray scale images at $10$ fps and its view point was directed towards the direction of the vehicle motion. 
The vehicle and camera poses are estimated by a RTK-GPS and they are the initial input of our system together with the video stream.

Among all the KITTI sequences we choose the 0095 (268 frames) and 0104 (313 frames) from the raw dataset and, sequences 03 (801 frames) and 04 (271 frames) from the odometry dataset. 
They depict four different urban scenarios: the 0095 shows a narrow environment where the building fa\c{c}ades are close to the camera, the 0104 captures a wide road, while the 03 and 04 sequences provide a varied landscape mixing natural (trees and bushes) and man-made (houses, cars) features.
We run the tests on a 4 Core i7-2630QM CPU at 2.2Ghz (6M Cache) with 6GB of DDR3 SDRAM. To have a qualitative overview of the results we refer the reader to the video in the supplementary material.

\begin{figure}[t]
  \centering
  \includegraphics[width=0.43\textwidth]{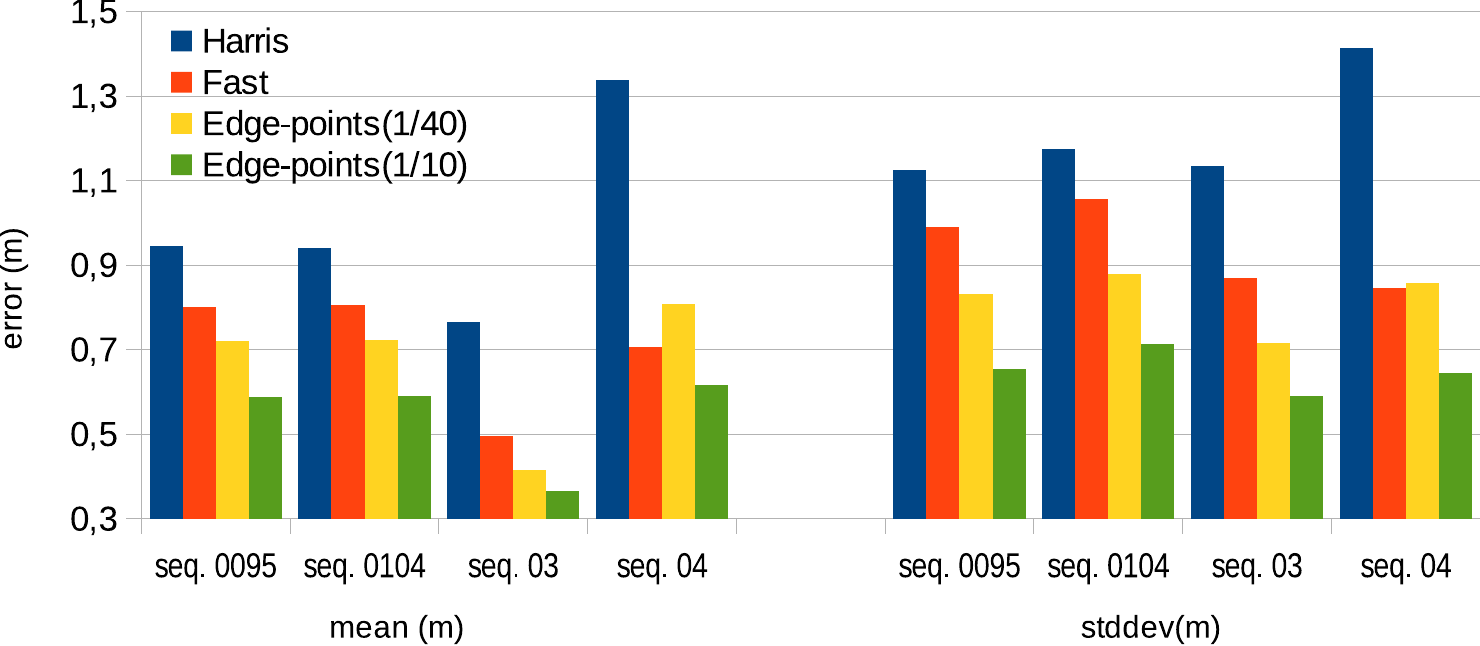}
  \caption{Reconstruction absolute errors of the proposed algorithm (Edge-Point) versus two classical feature based 3D reconstruction (Harris, FAST).}
   \label{tab:comp}
\end{figure}

\begin{figure}[t]
  \centering
  \includegraphics[width=0.4\textwidth,height=0.18\textwidth]{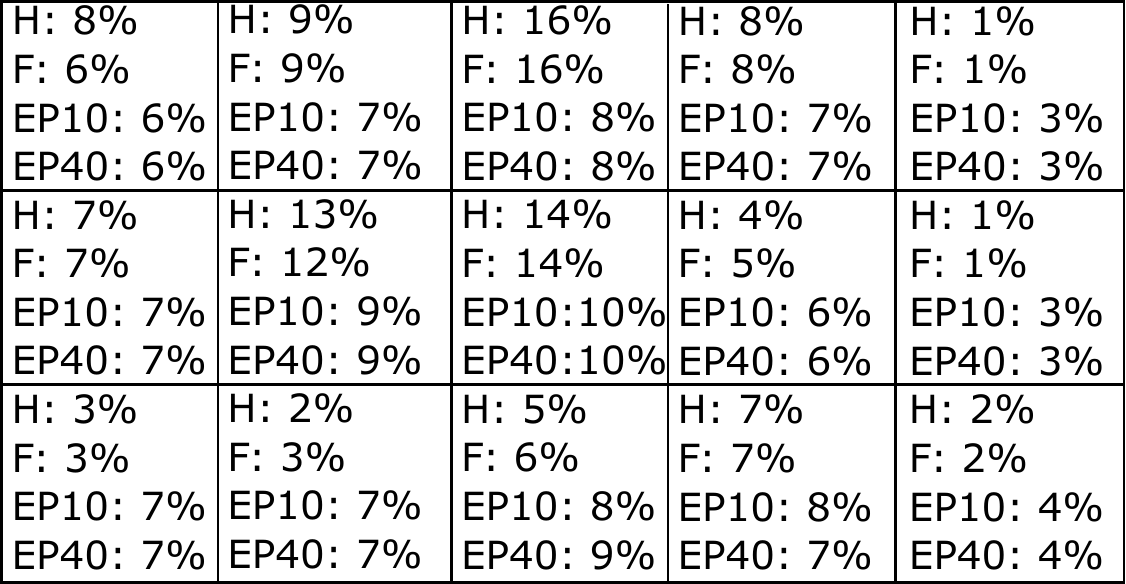}
  \caption{Distribution of Harris corner (H), FAST (F) and Edge-Points with downsampling $\frac{1}{10}$ (EP10) and $\frac{1}{40}$ (EP40) on the 0095 sequence.}
   \label{fig:distr}
\end{figure}

To provide a quantitative evaluation we compared the reconstructed meshes with the very accurate point clouds measured by the Velodyne of the KITTI dataset through the CloudCompare tool \cite{CloudCompare}.
This tool was used to compute the reconstruction error, i.e., the average of the distances between each Velodyne point and the nearest mesh triangle.
Fig. \ref{tab:comp} shows the comparison, on the same dataset, between the reconstruction with the proposed Edge-Points cloud and the reconstruction with the FAST and the Harris corner point clouds as in \cite{litvinov_Lhiuller14}.  
We adopted two different edge downsampling rate (low $\frac{1}{40}$ and high $\frac{1}{10}$) to verify that the accuracy gain was not a matter of number of reconstructed features, but it was due to the better choice. 
Indeed, Table \ref{tab:reconstrPt} shows that, even if the reconstructed Edge-Points with low downsampling rate are significantly less than the reconstructed points using classical features, the accuracy of the manifold estimated in the former case is always better with except to one case (sequence 04 with respect to FAST).  
The good fitting of the Edge-Points to the real 3D curves, lets the reconstructed surface to lay closer to the real one; this allows our Edge-Point reconstruction approach to outperform reconstructions upon non-Edge-Points.
Fig. \ref{fig:distr} shows how Edge-Points have a more homogeneous distribution on the images, with respect to the other features: we subdivided the images of the sequence into a 3x5 grid and we report the percentage of extracted features for each cell.

\begin{figure}[t]
  \centering
  \includegraphics[width=0.4\textwidth]{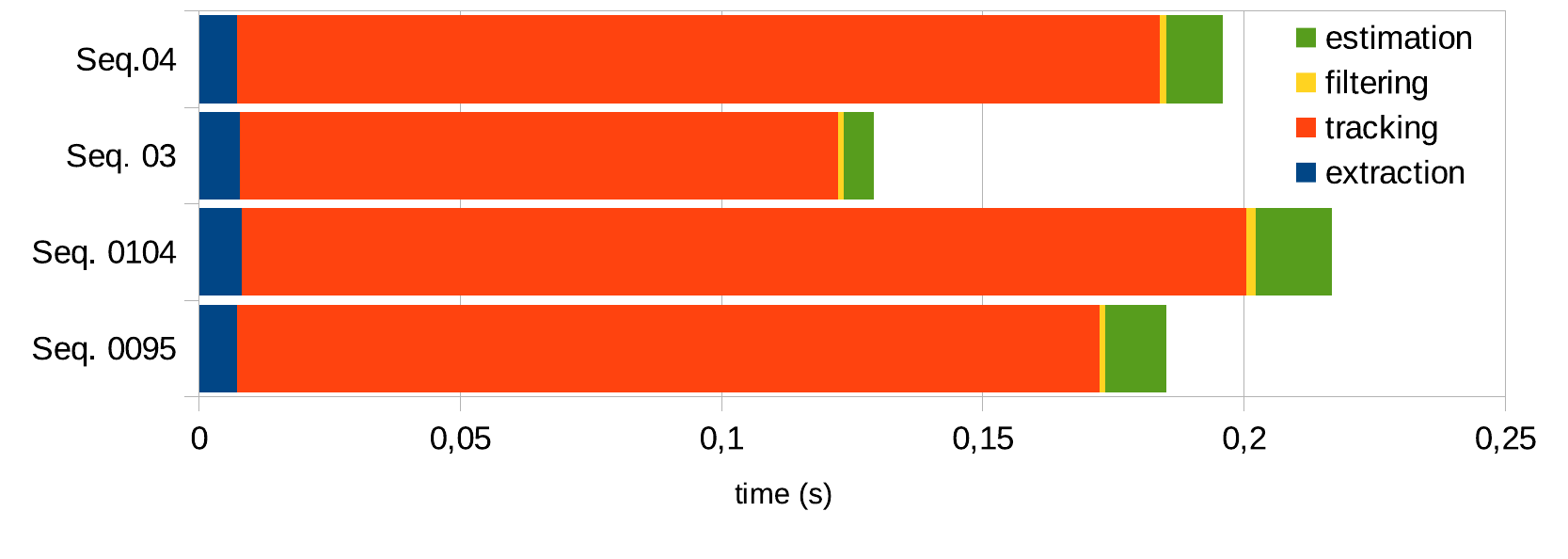}
  \caption{Per-frame time (in seconds) of Edge-Point estimation ($\frac{1}{10}$ downsampling rate).}
   \label{tab:timing}
\end{figure}

To understand how Edge-Points extraction, tracking, filtering and estimation affect the performance of our reconstruction algorithm we reported the timing in Fig. \ref{tab:timing} for Edge-Points with downsampling rate $\frac{1}{10}$.
In our experiments, the tracking, filtering and estimation processing times were proportional to the number of extracted features, while the extraction depends on the selected features: the mean per-frame times are 0.0044s (Harris),  0.0010s (FAST),  0.0036 (Edge-Points with $\frac{1}{40}$ downsampling),  0.0037 (Edge-Points with $\frac{1}{10}$ downsampling). FAST is the fastest feature to extract but the impact of this step on the overall 3D estimation pipeline is almost negligible (1\% to 3\% of the pipeline).

\begin{table}[t]
  \caption{Mean number of reconstructed (and extracted) points per keyframe.}
  \scriptsize
   \label{tab:reconstrPt}
   \centering
   \begin{tabular}{p{0.2\columnwidth}cccc}
   \toprule 
                                              & 0095            & 0104        &  03         & 04   \\
   \hline   
   {Harris}                                   & 423 (2556)      & 561 (2979)  & 946 (3342)  & 692 (2036) \\
   %\hline
   {FAST}                                     & 550 (3463)      & 865 (3950)  & 1215 (4358) & 953 (2850) \\
   %\hline
  {Edge-Point (${1}/{40}$ downs.)}   & 165 (1327)      & 267 (1485)  & 404 (1650)  & 382 (1277) \\
   %\hline
   {Edge-Point (${1}/{10}$ downs.)}  & 656 (5310)      & 946 (5938)  & 1615 (6598) & 1524 (5106)  \\
    \bottomrule
  \end{tabular}
  \end{table} 
  
% \begin{table}[t]
%   \caption{Mean number of reconstructed (and extracted) points per keyframe.}
%   \footnotesize
%    \label{tab:reconstrPt}
%    \centering
%    \begin{tabular}{p{0.25\columnwidth}cccc}
%    \toprule 
%                                               & 0095    & 0104    & 03      & 04   \\
%    \hline   
%    {Harris}                                   & 301     & 300 ()     & 112     & 151 \\
%    %\hline
%    {FAST}                                     & 1291    & 1322 (3950)    & 1732    & 1152 \\
%    %\hline
%   {Edge-Point (${1}/{40}$ downsample rate)}   & 447     & 521  (2979)   & 266     & 467 \\
%    %\hline
%    {Edge-Point (${1}/{10}$ downsample rate)}  & 1785    & 2077 (5938)   & 1062    &1858  \\
%     \bottomrule
%   \end{tabular}
%   \end{table} 
In Table \ref{tab:numArtifacts} we show the effect of the Inverse Cone Heuristic. We manually counted the visually critical artifacts in the mesh reconstructed with and without the heuristic; in parenthesis, we reported the number of artifacts affecting the camera traversability path. The heuristic diminished significantly the number of artifacts by $68$\% up to $85$\%, depending on the sequence considered. 
A fair comparison with the method in \cite{litvinov_Lhiuller14} is not possible, since their dataset and their code is not publicly available. We only point out that, in their experiments, they reported \cite{litvinov_Lhiuller14} an artifact removal rate of $35$\%.

\begin{figure}
\centering
\begin{tabular}{c}
\includegraphics[width=0.7\columnwidth]{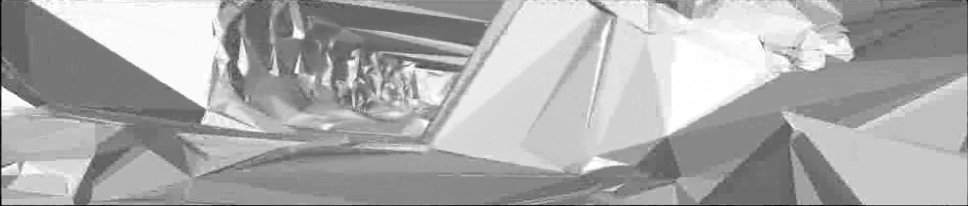}\\
(a)\\
\includegraphics[width=0.7\columnwidth]{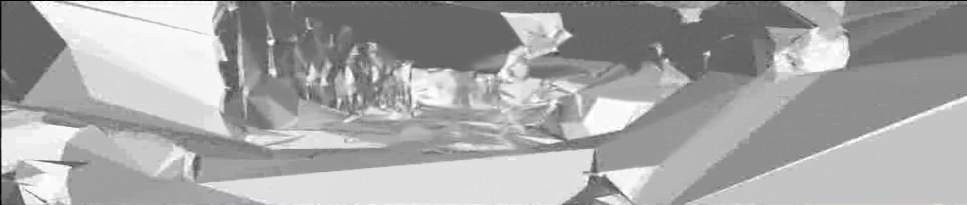}\\
(b) 
\end{tabular}
\caption{Example of preemptive artifact removal: (a) without  and (b) with the Inverse Cone Heuristic.}
\label{fig:exampleArt}
\end{figure}

We are also able to provide a qualitative comparison about how the Inverse Cone Heuristic affects the performance of the reconstruction. The method in \cite{litvinov_Lhiuller14} takes $0.43$s per frame on a Xeon W3530 at 2.8Ghz (8M Cache), whose performances are very similar to our machine.  Our datasets are different, but depict a similar urban scenario, and our approach (Table \ref{tab:inverseTiming}) runs one to two order of magnitude faster, thanks to the CGAL \cite{cgal} triangulation data structure which enable very efficient access to  tetrahedra neighboring the ones traversed by the camera-to-point viewing rays. 
Fig. \ref{fig:exampleArt} shows a mesh without and with the inverse cone heuristic; the big artifact occluding the camera trajectory in (a) disappears in (b).

% Finally, we are able to conclude that the Inverse Cone Heuristic is surely effective, moreover it is possible to apply it complementary to the algorithm presented in \cite{litvinov_Lhiuller14}, since it has a really low impact on the performances. 

%

%----------------------------------------------------------------------------------------------------------------------------
\section{Conclusion and future works}
\label{sec:conclusion}
In this paper we have shown that Edge-Points represent a very convenient choice to build a 3D Delaunay triangulation for the Space Carving reconstruction, especially in urban scenarios.
We have shown how to successfully reconstruct their 3D positions by tracking their successive projections in the video images and by filtering the results of the KLT tracker with simple constraints. On these reconstructed points we incrementally built a 3D triangulation to reconstruct a manifold surface with a novel version of the algorithm in \cite{Litvinov_Lhuillier_13} and \cite{litvinov_Lhiuller14} improved by means of the Inverse Cone Heuristic.

The results reached by our algorithm showed that in urban scenarios the Edge-Points estimation enables a detailed reconstruction, which is better then those obtained by  using only stable features, such as Harris or FAST corners.

To deal with visual artifacts affecting the reconstructed manifold, we proposed a very fast method to preemptively avoid their creation. In the experiments the manifolds reconstructed with this heuristic are almost clear from visual artifacts.
Future works involve the preemptive filtering of some of the Edge-Points not belonging to the real-world edges or laying in a very low parallax regions, and  the management of moving 3D points inside the Delaunay triangulation as in \cite{Romanoni15a}.

\begin{table}[t]
  \caption{Number of artifacts w/o and w/ the Inverse Cone Heuristic. In parenthesis number of artifacts affecting the camera traversability path.}
  \footnotesize
   \label{tab:numArtifacts}
   \centering
   \begin{tabular}{p{0.25\columnwidth}cccc}
   \toprule 
                               & 0095  & 0104& 03 & 04   \\
   \hline
%    {w/o ICH }                 & 21&21& 40& 22\\
%    w/  ICH                    & 4&3& 12& 7\\
%    \% removed                 & 80&85& 70& 68\\
   w/o ICH                 & 21 (4) &21(2)& 40(15)& 22(7)\\
   w/  ICH                    & 4(0) &3(1)& 12(3)& 7(1)\\
 \% removed                 & 80(100) &85(50)& 70(80)& 68(85)\\
    \bottomrule
  \end{tabular}
  \end{table}

\begin{table}[t]
  \caption{Per-frame time (in seconds) of the Inverse Cone Heuristic (ICH) for preemptive artifacts removal.}
  \footnotesize
   \label{tab:inverseTiming}
   \centering
   \begin{tabular}{cccc}
   \toprule  
     0095  & 0104&03 & 04     \\
   \hline
     0.002 & 0.003& 0.010 & 0.001  \\
    \bottomrule
  \end{tabular}
  \end{table} 

\section*{Acknowledgements}
Work partially funded by the SINOPIAE project, from the Italian Ministry of University and Research and Regione Lombardia, and by MEP (Maps for Easy Paths) project from the Politecnico di Milano under the POLISOCIAL program.

\bibliographystyle{IEEEtran}
\bibliography{IEEEabrv,biblioEdgePoint}

\end{document}